\documentclass{INTERSPEECH2023}

\usepackage{microtype}
\usepackage{subcaption}
\usepackage{graphicx}

\interspeechcameraready

\newcommand{\model}{\textsc{segue}}

\title{Sentence Embedder Guided Utterance Encoder (\model{}) for Spoken~Language~Understanding}
\name{Yi Xuan Tan, Navonil Majumder, Soujanya Poria}
\address{
  Singapore University of Technology and Design, Singapore
}
\email{yixuan\_tan@sutd.edu.sg, navonil\_majumder@sutd.edu.sg, sporia@sutd.edu.sg}

\begin{document}

\maketitle
 
\begin{abstract}
The pre-trained speech encoder wav2vec 2.0 performs very well on various spoken language understanding (SLU) tasks. However, on many tasks, it trails behind text encoders with textual input. To improve the understanding capability of SLU encoders, various studies have used knowledge distillation to transfer knowledge from natural language understanding (NLU) encoders. We use a very simple method of distilling from a textual sentence embedder directly into wav2vec 2.0 as pre-training, utilizing paired audio-text datasets. We observed that this method is indeed capable of improving SLU task performance in fine-tuned settings, as well as full-data and few-shot transfer on a frozen encoder. However, the model performs worse on certain tasks highlighting the strengths and weaknesses of our approach.
\end{abstract}

\section{Introduction}

Spoken language understanding (SLU) tasks~\cite{mehrish2023review} have greatly benefited from modern transformer-based speech encoders, such as wav2vec 2.0 \cite{Baevski20a}, to the point that end-to-end models can now replace cascaded models, doing away with the ASR component. However, these models still trail behind equivalent tasks in the textual modality \cite{Borgholt21a}. This could be interpreted as that textual models still contain much knowledge that speech models lack.

One way to transfer knowledge from one model to another is through knowledge distillation (KD) \cite{Hinton15a}. KD was originally used to transfer knowledge from larger models to smaller ones. However, with the assumption that parallel speech and text inputs are roughly equivalent, one could perform cross-modal KD from textual models to speech models. This idea has seen success in previous studies \cite{Cho20a, Denisov20a, Kim21a}.

In this paper, we perform KD directly from a sentence embedder to a wav2vec 2.0 speech encoder. This produces a pre-trained sentence-embedder-guided utterance encoder (\model{}), which can be used for sequence-level SLU tasks. We conduct experiments on SLU tasks---sentiment and emotion detection on speech modality, fluent speech commands (FSC), and automatic speech recognition (ASR). Our results demonstrate that \model{} is capable of improving performance over vanilla wav2vec 2.0, to varying degrees. We show results for fine-tuned settings, as well as full-data and few-shot transfer with a frozen encoder. However, we also observe that \model{} performs worse on two of said tasks. Our code is available on GitHub\footnote{https://github.com/declare-lab/segue}.

\section{Related Work} 

Our approach aims to produce an utterance embedding, which was inspired by sentence encoding problems in the literature. In particular, we base the overall concept on Sentence-BERT \cite{Reimers19a}. We adopt the idea of knowledge distillation (KD) proposed by Hinton et al., and textual-to-spoken cross-modal KD has previously been explored by the following works. Cho et al. \cite{Cho20a} perform KD directly with the downstream task, whereas we attempt to perform KD as pre-training for different downstream tasks to reuse. Denisov et al. \cite{Denisov20a} perform KD in pre-training as we do, but they construct an utterance encoder by initializing from a trained ASR model's backbone connected to a trained NLU backbone. In contrast, we attempt to distill knowledge directly into a wav2vec 2.0 encoder without ASR training and without a trained NLU module on top. Kim et al. \cite{Kim21a} use a more complex architecture and perform KD in both pre-training and fine-tuning stages.

\section{Method}

\begin{figure}[t]
  \centering
  \includegraphics[width=\linewidth]{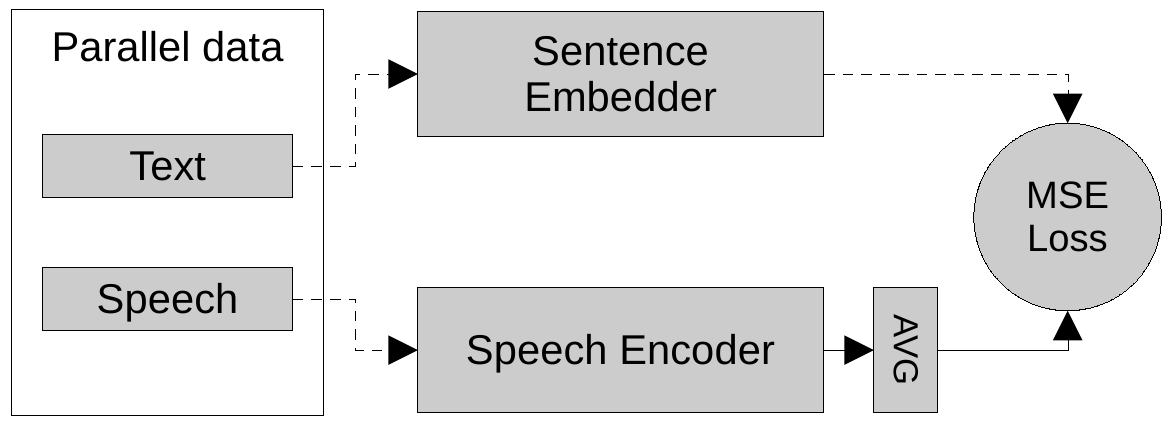}
  \caption{Diagram of \model{} pre-training. Solid arrows indicate gradient flow in the backward pass, and dashed arrows indicate the lack thereof.}
  \label{fig:method-diagram}
\end{figure}

Our method is based on the simple idea of distilling knowledge from a sentence embedder $T$ into a speech encoder $S$, with the assumption that the textual input $t$ and speech input $s$ are roughly equivalent in meaning, as shown in \autoref{fig:method-diagram}. To produce a single fixed-length embedding vector given a speech input $s$ of length $l$, the sequence of speech encoder output vectors $[S_1(s), S_2(s), \dots, S_l(s)]$ is average-pooled into a single embedding vector $S(s)$, similar to Sentence-BERT \cite{Reimers19a}:
\begin{align}
    S(s) = \frac{1}{l} \sum^l_i S_i(s)
\end{align}
A mean squared error (MSE) loss is then used to align the outputs of the speech student model to that of the textual teacher model.
\begin{align}
    L(s, t) = \lVert S(s) - T(t) \rVert_2^2
\end{align}

During pre-training, the loss $L(s, t)$ is computed, and its gradient w.r.t. the parameters of speech encoder $S$ is computed to train said encoder, while sentence embedder $T$ remains frozen.

We used the all-mpnet-v2-base checkpoint provided by the sentence-transformers package \cite{Reimers19a} as the sentence embedder. The speech encoder is a pre-trained wav2vec 2.0 base encoder as released by Baevski et al. \cite{Baevski20a}. Hence, both models produce a 768-dimensional embedding vector. Our model has 95 million parameters, equal to that of the wav2vec 2.0 base model.

This method requires text and speech of approximately equivalent meaning, so we used the 960 hours LibriSpeech dataset \cite{Panayotov15a} as a source of parallel text and speech data.

One potential weakness of this method is that this may not capture the rich paralinguistic features in speech such as prosody, which are typically thought to be important for the semantic content of utterances. However, there may still be enough information in the training data for the model to learn to make use of these features to some extent.

\subsection{Setting details}

For pre-training, we used a linear LR schedule with warmup over the first 5000 steps, and a peak learning rate of 3e-5. The AdamW optimizer \cite{Loshchilov19a} was used with beta parameters 0.9 and 0.999, and 0 weight decay. We trained for 10 epochs on the LibriSpeech dataset. We used two A6000 GPUs with a per-device batch size of 8, totaling 16. The training took approximately 28 hours. We saved a checkpoint every 5000 steps, and average the parameters of the last 10 checkpoints to produce the final model.

\subsection{Contrastive pre-training}

Inspired by CLIP \cite{Radford21a}, our initial idea was to use the InfoNCE loss \cite{Oord18a} to contrastively train a speech encoder and a text encoder to share the same embedding space. The task was set up such that the model has to predict, within a batch, which pairs of textual sentences and utterances correspond to each other, i.e. positive pairs. We monitored the loss, as well as the mean positive-pair similarity:
\begin{align}
    S_{+}(U, S) = \frac{1}{|U|} \sum^{|U|}_{i=1} S(u_i, s_j)
\end{align}
where $u \in U$ is a batch of utterances, $s \in S$ is a batch of textual sentences s.t. $u_i$ and $j_i$ form positive pairs, and $S(u_i, s_j)$ is the cosine similarity between the $i$th utterance embedding and the $j$th sentence embedding.

We found that this quickly causes catastrophic forgetting in the text encoder, and the speech encoder was not able to learn effectively from the setting. Therefore, we froze the text encoder, at which point it resembled a KD setting, but using InfoNCE rather than MSE loss, along with a learnable temperature for use with InfoNCE. We found that $S_+(U, S)$ correlated with good downstream task performance, but the loss did not. However, when we explored using KD with an MSE loss, it matched the best contrastive models that we had, and checkpoint selection with loss in KD was more consistent than with $S_+(U, S)$ in InfoNCE. Hence, we decided to use KD instead of contrastive training.

\section{Experiments}

We perform pre-training of \model{} using the above method, and then evaluate the model on several downstream tasks, in fine-tuned settings (i.e. with a tunable encoder), as well as full-data and few-shot transfer settings with a frozen encoder (henceforth referred to simply as "full-data transfer" and "few-shot transfer"). We compare the results with the vanilla wav2vec 2.0 baseline. These tasks were done with the addition of a single linear layer on top of the encoder. We use the same AdamW setting as in pre-training, and a linear LR schedule with warmup unless otherwise stated. As both models were trained on \SI{16}{\kilo\hertz} mono audio, any audio input not in that format was converted into said format before being fed into the models.

\subsection{Downstream tasks}

We evaluate on sentiment regression with MOSEI \cite{Zadeh18a}, sentiment and emotion classification with MELD \cite{Poria19a}, intent classification with the MInDS-14 \cite{Gerz21a} en-US subset, and intent classification and slot-filling with Fluent Speech Commands (FSC) \cite{Lugosch19a}. Additionally, we evaluated on ASR with FLEURS \cite{Conneau22a} as it may help highlight the weaknesses of \model{}.

For MInDS-14, we randomly split the data into train-development-test subsets in a 60:20:20 ratio.

For multimodal tasks, i.e. MELD and MOSEI, we used only the speech modality, so we expected that the results would be far from state-of-the-art where the textual modality and potentially visual modality are used as well.

It should be noted that our copy of MOSEI has some missing videos, and due to the raw dataset no longer being publicly available, our results may not be fully comparable with results from other works.

\section{Results}

\begin{table*}[t]
    \caption{Results of MOSEI fine-tuning. The number after $\pm{}$ indicates standard deviation.}
    \label{tab:mosei}
    \centering
    \begin{tabular}{ l c c c c c}
        \toprule
        Model & MAE & Corr & F1 & Acc2 & Acc7 \\
        \midrule
        \multicolumn{6}{c}{Fine-tuning} \\
        \midrule
         w2v 2.0 & .799 $\pm$ .006 & .458 $\pm$ .016 & 72.9 $\pm$ 0.4 & 73.4 $\pm$ 0.2 & 37.5 $\pm$ 0.3 \\
         (averaged) & .799 $\pm$ .002 & .460 $\pm$ .009 & 72.7 $\pm$ 0.4 & 73.0 $\pm$ 0.4 & 39.2 $\pm$ 0.9 \\
         \model{} & \bfseries .781 $\pm$ .006 & \bfseries .473 $\pm$ .015 & 73.1 $\pm$ 0.8 & 73.5 $\pm$ 0.3 & \bfseries 40.5 $\pm$ 0.5 \\
         (averaged) & .797 $\pm$ .008 & \bfseries .473 $\pm$ .006 & \bfseries 73.5 $\pm$ 0.5 & \bfseries 73.6 $\pm$ 0.5 & 40.1 $\pm$ 1.5 \\
        \midrule
        \multicolumn{6}{c}{Full-data transfer} \\
        \midrule
         w2v 2.0 & .855 & .246 & 63.3 & 63.5 & 39.8 \\
         \model{} & \bfseries .836 & \bfseries .326 & \bfseries 65.5 & \bfseries 65.4 & \bfseries 40.5 \\
        \bottomrule
    \end{tabular}
\end{table*}

\begin{figure*}[t]
\centering
\begin{subfigure}[t]{0.33\textwidth}
  \includegraphics[width=\linewidth]{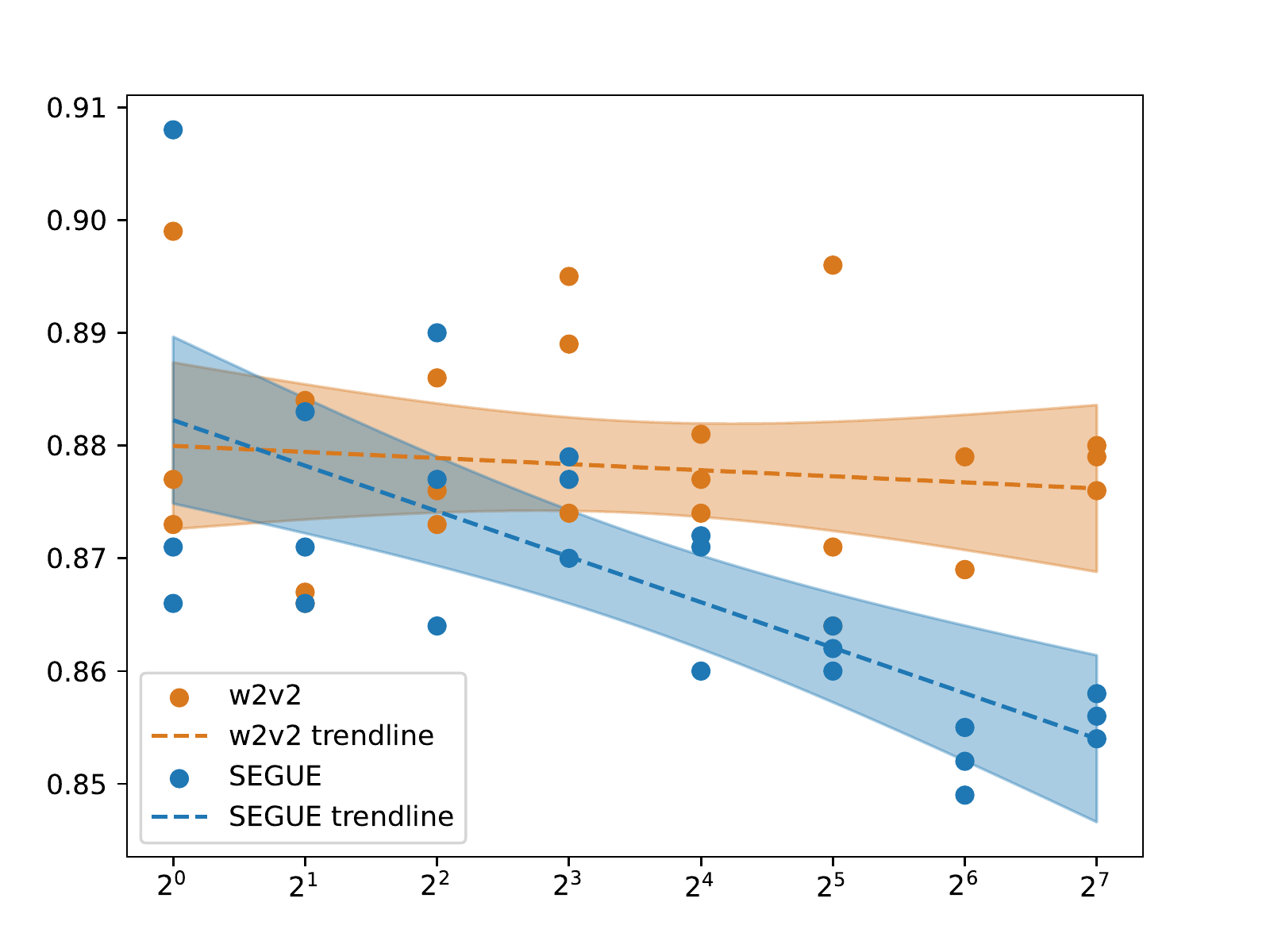}
  \caption{mean absolute error (MAE)}
  \label{fig:mosei-few-shot-mae}
\end{subfigure}%
\begin{subfigure}[t]{0.33\textwidth}
  \includegraphics[width=\linewidth]{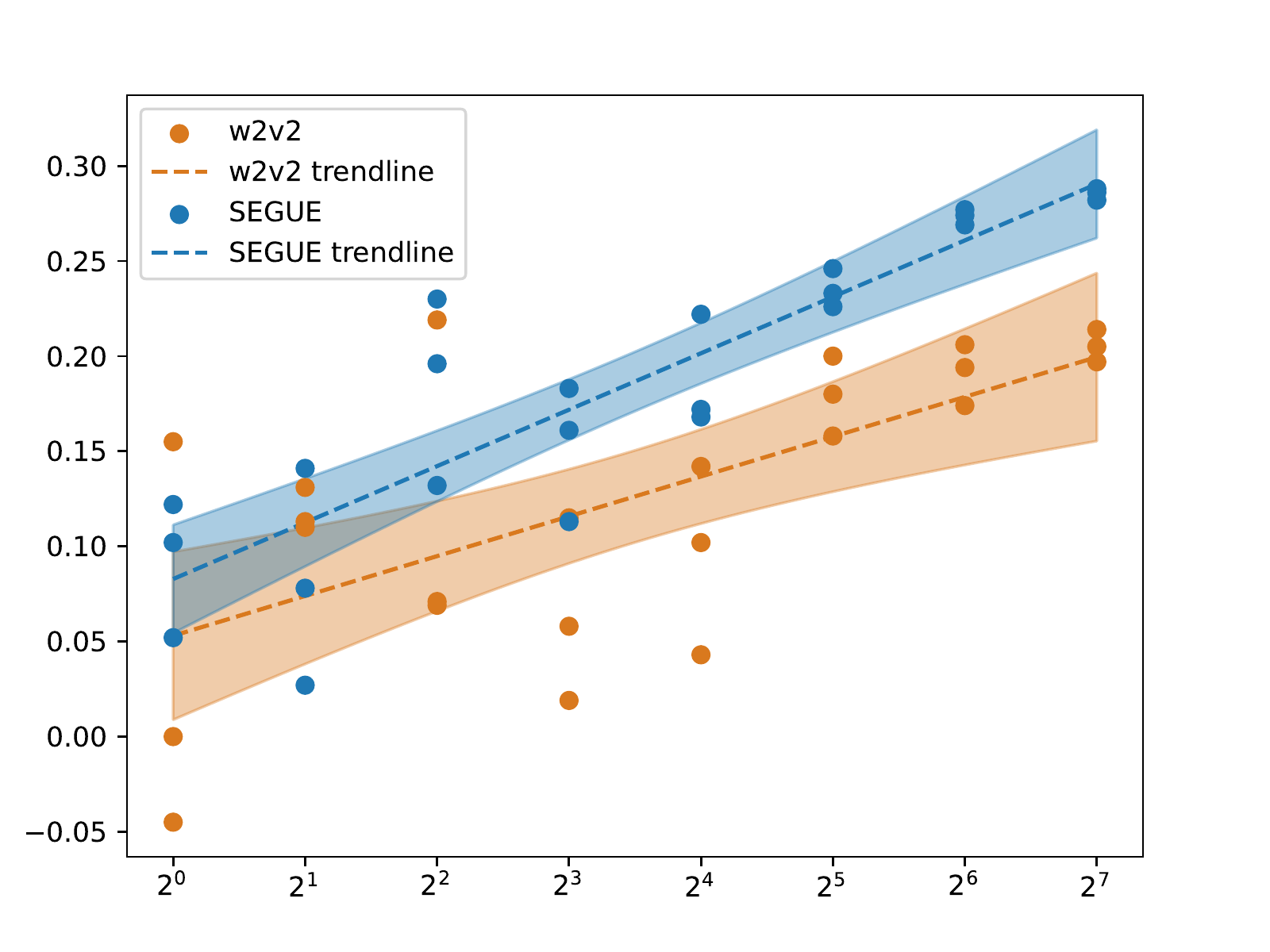}
  \caption{Pearson correlation coefficient}
  \label{fig:mosei-few-shot-corr}
\end{subfigure}%
\begin{subfigure}[t]{0.33\textwidth}
  \includegraphics[width=\linewidth]{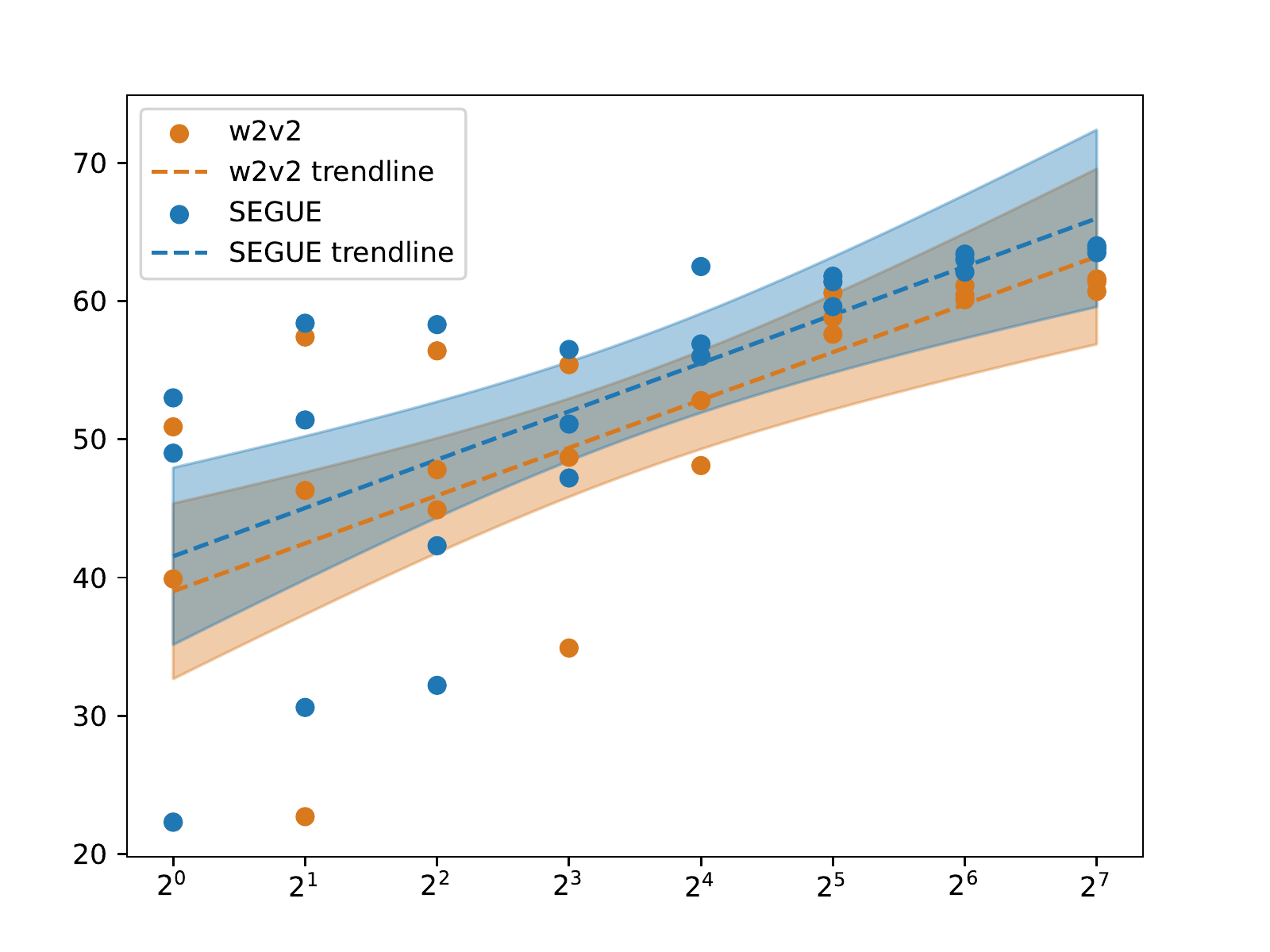}
  \caption{weighted F1 score}
  \label{fig:mosei-few-shot-f1}
\end{subfigure}
\caption{Plot of MOSEI k-shot per class performance against k.}
\end{figure*}

\subsection{MOSEI}

For the MOSEI sentiment task, the model was trained with a regression task, and rounding was used to retrofit the task as a classification task during evaluation. The fine-tuning and full-data transfer results are reported in \autoref{tab:mosei}. For full-data and few-shot transfer, we use an analytic solver for ridge regression with $\alpha = 100$.

For fine-tuning, we trained \model{} for 3 epochs with 30\% warmup steps, but we found that vanilla wav2vec 2.0 needed more epochs to converge so the baseline was trained for 5 epochs with 20\% warmup steps. We trained with a peak LR of 3e-5 and a batch size of 8. We conducted three runs and report the mean and standard deviation of the metrics. We tried averaging the last 10 checkpoints, which we found tend to boosted overall performance but worsen MAE. Hence, we report both best-checkpoint and averaged-checkpoint results.

For both fine-tuning and full-data transfer, \model{} improves performance across all metrics.

For few-shot transfer, we trained with $k$ samples per class, under the seven-class setup, obtained by rounding the labels to the nearest integers. Due to the size of the data table, we instead report the results as plots -- \autoref{fig:mosei-few-shot-mae}, \autoref{fig:mosei-few-shot-corr}, and \autoref{fig:mosei-few-shot-f1}. In terms of MAE and correlation, \model{} learns faster than vanilla wav2vec 2.0. In terms of F1 score, \model{} has a head start and maintains approximately the same lead as $k$ grows.

\begin{table}[t]
    \caption{Results of MELD fine-tuning and full-data transfer. The number after $\pm{}$ indicates standard deviation.}
    \label{tab:meld}
    \centering
    \begin{tabular}{l c c}
        \toprule
        Model & Sentiment F1 & Emotion F1 \\
        \midrule
        \multicolumn{3}{c}{Fine-tuning (LR = 3e-6)} \\
        \midrule
         w2v 2.0 & 46.8 & 37.2 \\
         (averaged) & 47.3 & 39.3 \\
         \model{} & \bfseries 53.2 & 41.0\\
         (averaged) & \bfseries 53.2 & \bfseries 41.1 \\
        \midrule
        \multicolumn{3}{c}{Fine-tuning (LR = 3e-5)} \\
        \midrule
         \model{} & 53.3 & 42.3 \\
         (averaged) & \bfseries 54.1 & \bfseries 47.2 \\
        \midrule
        \multicolumn{3}{c}{Full-data transfer} \\
        \midrule
         w2v 2.0 & 45.0 $\pm$ 0.7 & 34.3 $\pm$ 1.2 \\
         \model{} & \bfseries 45.8 $\pm$ 0.1 & \bfseries 35.7 $\pm$ 0.3 \\
        \bottomrule
    \end{tabular}
\end{table}

\begin{figure}[t]
  \centering
  \includegraphics[width=\linewidth,trim=0 0.5cm 0 0.5cm,clip]{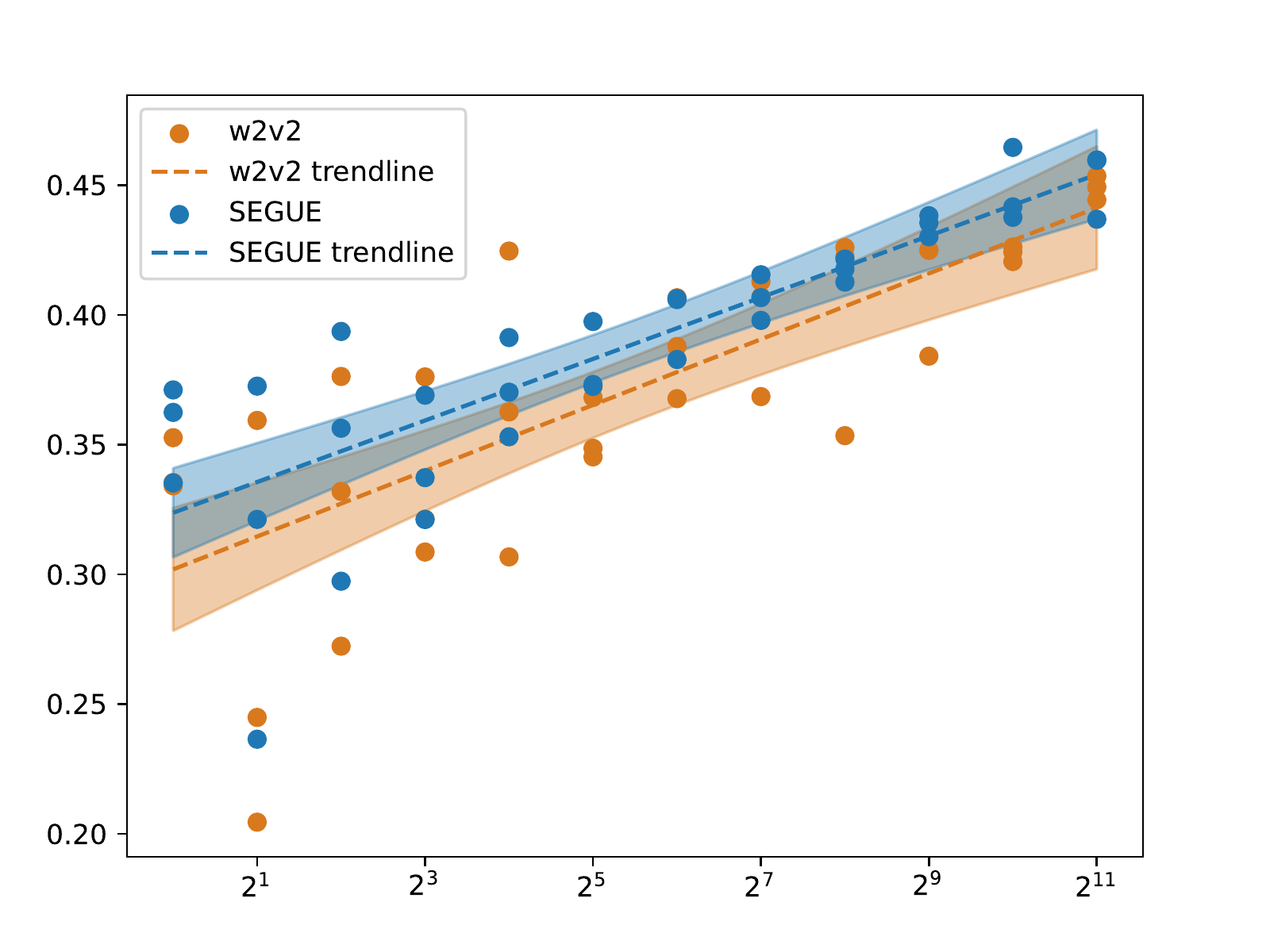}
  \caption{Plot of MELD k-shot per class sentiment F1 score against k.}
  \label{fig:meld-few-shot-sent}
\end{figure}

\begin{figure}[t]
  \centering
  \includegraphics[width=\linewidth,trim=0 0.5cm 0 0.5cm,clip]{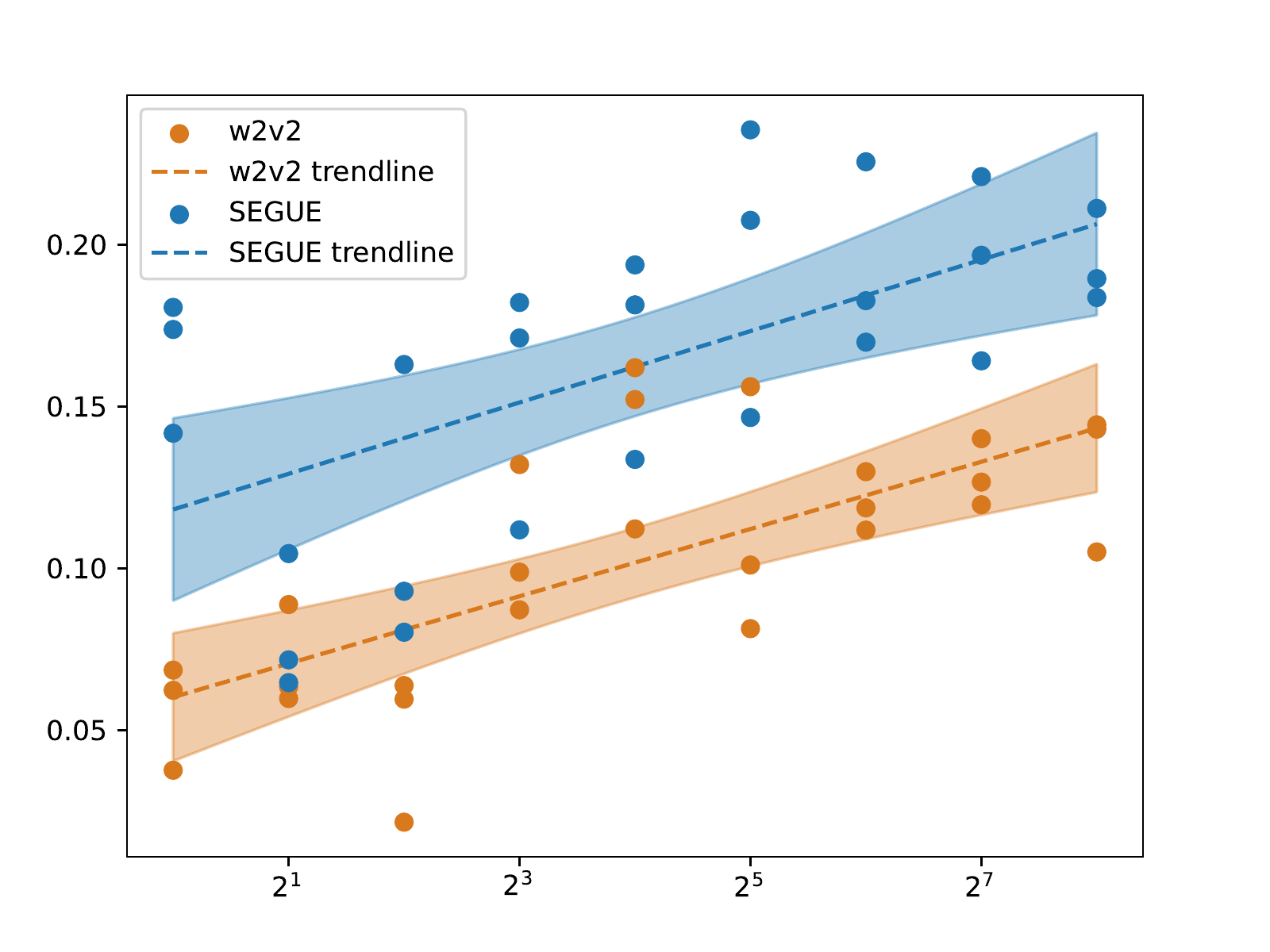}
  \caption{Plot of MELD k-shot per class emotion F1 score against k.}
  \label{fig:meld-few-shot-emo}
\end{figure}

\subsection{MELD}

The results for fine-tuning and full-data transfer for MELD are shown in \autoref{tab:meld}. \model{} outperforms the baseline in both settings.

For fine-tuning, we trained for 5 epochs with 20\% warmup steps and a batch size of 8. The wav2vec 2.0 baseline could not converge with a learning rate of 3e-5 which we used for \model{}, so we fine-tuned it at 3e-6 instead, and evaluated \model{} in that setting as well. We also average the last 10 checkpoints and report the performance.

For full-data transfer, we trained for 20 epochs at a peak LR of 1e-3, with 10\% warmup steps, and a batch size of 8. Due to the relatively close results and the training being somewhat noisy, we once again run thrice and report the average and standard deviation.

For few-shot transfer, we trained for 5 epochs at a peak LR of 1e-2 with 10\% warmup steps. The results are once again presented as plots, in \autoref{fig:meld-few-shot-sent} and \autoref{fig:meld-few-shot-emo}. The plots show that \model{} has a head start, and mostly maintains that advantage as $k$ grows.

\begin{table}[t]
    \caption{Results of FSC fine-tuning and full-data transfer. The number after $\pm{}$ indicates standard deviation.}
    \label{tab:fsc}
    \centering
    \begin{tabular}{l c}
        \toprule
        Model & Exact match accuracy \\
        \midrule
        \multicolumn{2}{c}{Fine-tuning} \\
        \midrule
         w2v 2.0 & \bfseries 99.6 $\pm$ 0.0 \\
         \model{} & \bfseries 99.6 $\pm$ 0.1 \\
        \midrule
        \multicolumn{2}{c}{Full-data transfer} \\
        \midrule
         w2v 2.0 & \bfseries 94.7 \\
         \model{} & 90.6 \\
        \bottomrule
    \end{tabular}
\end{table}

\subsection{Fluent Speech Commands}

As with other fine-tuning settings where the results are close, we ran each experiment thrice. For fine-tuning, we trained for 5 epochs at a peak LR of 3e-5, with 10\% warmup steps and a batch size of 8. For full-data transfer, we trained for 60 epochs at a peak LR of 1e-2, with 10\% warmup steps and a batch size of 8.

As can be seen in \autoref{tab:fsc}, wav2vec 2.0 achieves approximately the same accuracy as \model{} on fine-tuning. More surprisingly, wav2vec 2.0 is able to achieve 94.7\% accuracy on full-data transfer even on a frozen encoder. In doing so it also outperforms \model{} which only managed to reach 90.6\% accuracy. We hypothesize that this is because the task can be performed well with less understanding capability, relying more on word detection capability. The relative ease of this task as shown by the high accuracy scores may also point toward this hypothesis. Since \model{} is trained against semantic embeddings alone, it is likely that it has lost some of the word detection capability of the original wav2vec 2.0 encoder, hence performing worse with a frozen backbone. However, it may not have completely forgotten the lost capability, hence it was still able to recover to the same performance as wav2vec 2.0 when fine-tuned.

Note that the above hypothesis may not apply to MInDS-14 despite it also being an intent classification task, as it contains utterances that are more free-form, unlike FSC’s mostly fixed sentence structure, and thus may benefit more from deeper reasoning.

\begin{table}[t]
    \caption{Results of MInDS-14 fine-tuning and full-data transfer. The number after $\pm{}$ indicates standard deviation.}
    \label{tab:minds14}
    \centering
    \begin{tabular}{l c c}
        \toprule
        Model & \# runs & Accuracy \\
        \midrule
        \multicolumn{3}{c}{Fine-tuning} \\
        \midrule
         w2v2 (all runs) & 6 & 46.5 $\pm$ 47.0 \\
         w2v2 (failed runs) & 3 & 3.5 $\pm$ 0.0 \\
         w2v2 (converged runs) & 3 & 89.4 $\pm$ 2.3 \\
         \model{} & 3 & \bfseries 97.6 $\pm$ 0.5 \\
        \midrule
        \multicolumn{3}{c}{Full-data transfer} \\
        \midrule
         w2v 2.0 & -- & 54.0 \\
         \model{} & -- & \bfseries 77.9 \\
        \bottomrule
    \end{tabular}
\end{table}

\begin{figure}[t]
  \centering
  \includegraphics[width=0.5\textwidth,trim=0 0.5cm 0 0.5cm,clip]{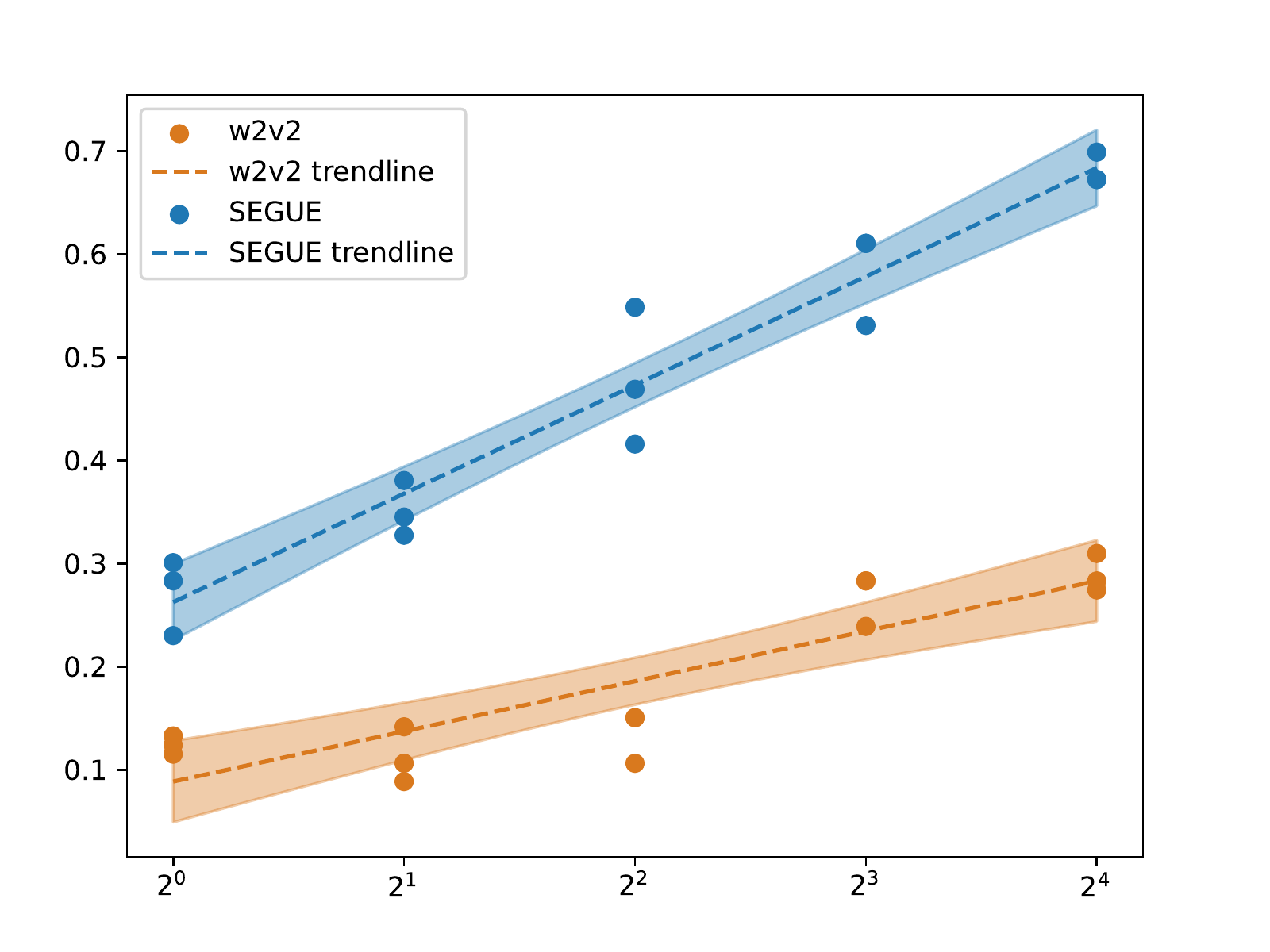}
  \caption{Plot of MInDS-14 k-shot per class accuracy against k.}
  \label{fig:minds14-few-shot}
\end{figure}

\subsection{MInDS-14}

The fine-tuning and full-data transfer results for MInDS-14 are shown in \autoref{tab:minds14}, and the few-shot transfer results in \autoref{fig:minds14-few-shot}.

For fine-tuning, we trained \model{} for 40 epochs at a peak LR of 3e-5, with 10\% warmup steps and a batch size of 16. The baseline once again needed longer to converge, so we trained for 60 epochs instead. We observed that the baseline could not stably converge, so we conducted 6 runs and broke down the results. On the other hand, \model{} managed to stabilize the training, and improved the converged performance as well.

For both full-data and few-shot transfer, \model{} drastically improves over the baseline.

\subsection{FLEURS ASR}

For FLEURS ASR, wav2vec 2.0 achieved a word error rate of 18.4, while \model{} achieved 23.1. This is as expected, as \model{} was trained as a sequence-level encoder, so it would likely not perform as well on ASR based on a connectionist temporal classification (CTC) \cite{Graves06a} loss, which is a token-level classification task. On top of that, if the aforementioned hypothesis regarding the relatively poor FSC performance is true, since ASR relies heavily on word detection capabilities it follows that \model{} would perform worse.

\section{Conclusion}

We produced \model{}, a pre-trained model for sequence-level SLU tasks. Our experiments demonstrated that it improves over the base wav2vec 2.0 model for many SLU tasks, some to a drastic degree, but performs worse on some tasks. We suggest that \model{}'s strength over wav2vec 2.0 is a deeper ability to understand speech, but it comes at a cost of some word detection capability. Hence, \model{} may perform worse on tasks that rely less on understanding and more on word detection, especially when the encoder is frozen, though some of said capability may be recovered in fine-tuning. Potential future work may include using a more diverse pre-training dataset to achieve more comprehensive KD, and possibly using multilingual sentence embedders to train a multilingual utterance encoder.

\clearpage

\bibliographystyle{IEEEtran}
\bibliography{mybib}

\end{document}